\documentclass[10pt,letterpaper]{article}
\usepackage[top=0.85in,left=2.75in,footskip=0.75in,marginparwidth=2in]{geometry}

\usepackage[utf8]{inputenc}

\usepackage{cite}

\usepackage{nameref,hyperref}

\usepackage{microtype}
\DisableLigatures[f]{encoding = *, family = * }

\raggedright
\usepackage{changepage}

\usepackage[aboveskip=1pt,labelfont=bf,labelsep=period,singlelinecheck=off]{caption}

\usepackage{lastpage,fancyhdr,graphicx}
\usepackage{epstopdf}
\fancyheadoffset[L]{2.25in}
\fancyfootoffset[L]{2.25in}

\usepackage{color}

\definecolor{Gray}{gray}{.25}

\usepackage{graphicx}
\usepackage{sidecap}

\usepackage{wrapfig}
\usepackage[pscoord]{eso-pic}
\usepackage[fulladjust]{marginnote}
\reversemarginpar

\usepackage{subfigure}
\usepackage{multirow}

\begin{document}
\vspace*{0.35in}

\begin{flushleft}
{\Large
\textbf\newline{Converting a Common Document Scanner to a Multispectral Scanner}
}
\newline
\\
Zohaib~Khan\textsuperscript{1,2},
Faisal~Shafait\textsuperscript{1,3},
Ajmal~Mian\textsuperscript{1,*}
\\
\bigskip
\bf{1} Department of Computer Science and Software Engineering, The University of Western Australia, Crawley, WA, Australia
\\
\bf{2} School of Information Technology and Mathematical Sciences, University of South Australia, Adelaide, SA, Australia
\\
\bf{3} School of Electrical Engineering and Computer Science (SEECS), National University of Science and Technology (NUST), Islamabad, Pakistan
\\
\bigskip
* ajmal.mian@uwa.edu.au

\end{flushleft}

\section*{Abstract}
We propose the construction of a prototype scanner designed to capture multispectral images of documents. A standard sheet-feed scanner is modified by disconnecting its internal light source and connecting an external multispectral light source comprising of narrow band light emitting diodes (LED). A document is scanned by illuminating the scanner light guide successively with different LEDs and capturing a scan of the document. The system is portable and can be used for potential applications in verification of questioned documents, cheques, receipts and bank notes.


\section{Introduction}

Forensic analysis of questioned documents involves a broad range of activities~\cite{leedham2003survey}. This includes establishing, whether a document originated from a particular source, backdated, forged or willfully manipulated. Disputes over the authenticity of bank cheques, purchase receipts, currency notes~\cite{baek2018detection} or seals in agreements~\cite{lee2012forged} can involve overwhelmingly complex legal procedures in resolution. In other cases, verification of genuineness of document source (written or printed) is also of significant interest to fraud detection~\cite{saini2016forensic}. The estimated age of a testament (will) can sometimes play a crucial role in the resolution of inheritance.

Traditionally, forensic scientists make empirical or experimental observations about a suspicious portion of the document in a forensic laboratory. The observations are then coupled with expert opinions to be presentable in a court-of-law. As this process largely relies on individual expertise and analysis, its consequences may be critical to the rights of a person, business or an organization. There is an interest in mechanisms for pre-examination of questioned documents before legally pursuing and bearing substantial costs in a court-of-law. Computerized forensic analysis has recently paved way for automatic document forgery detection using \emph{multispectral imaging}~\cite{khan2018automated,khan2018deep}. Multispectral or hyperspectral document scanners are generally comprised of bulky apparatus and require specialized laboratory environment for operation. This opens the need for the development of a portable multispectral document scanning system.

There are different ways of capturing multispectral images of a scene, each suits to a target application~\cite{shippert2003introduction}. A \emph{spatial scanner} simultaneously captures $(x,\lambda)$ dimensions of a scene, whereas the $y$ dimension is captured by movement of the sensor or the scene. It is suitable for scenarios where either the scene or the sensing platform is moving such as in remote sensing. A \emph{spectral scanner} simultaneously captures $(x,y)$ dimensions of a scene, whereas the $\lambda$ dimension is captured by spectral tuning~\cite{gat2000imaging}. It is specifically useful in a setup where both the scene and the sensing platform are stationary. In contrast, a \emph{snapshot spatio-spectral sensor} simultaneously captures $(x,y,\lambda)$ dimensions of a scene without the need for scanning. This method can effectively be used in conditions where the scene and the sensing platform are simultaneously moving. However, its complex sensor design incurs heavy costs limiting its use in applications like in-vivo imaging of organisms~\cite{hendargo2015snap}.

Previously, we proposed a spectral scanning system for capturing multispectral images of a document~\cite{khan2015automatic,khan2013hyperspectral}. Despite the simplicity of a static scene and sensor, the system was prone to artifacts of camera captured imaging (illumination, perspective etc.)~\cite{khan2013challenges}. In this work, we propose a \emph{spatio-spectral scanner} for capturing multispectral image of a document using a sheet-feed scanner, thus avoiding problems associated with camera capture. It captures one spatial dimension $x$ whereas the ($y,\lambda$) dimensions are sequentially captured by feeding the document and tuning illumination spectrum, respectively. In the following section, we describe the proposed multispectral document scanning system, in terms of its electrical, spectral, optical and thermal design.

\section{Multispectral Scanner Design}

The main components of the proposed multispectral document scanner are a standard document scanner and an external multispectral light source.

\begin{wrapfigure}{l}{0.3\linewidth}
\includegraphics[width=1\linewidth]{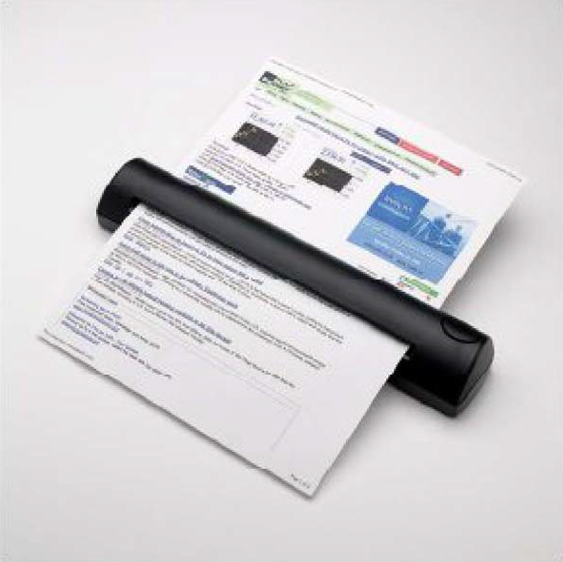}
\captionsetup{labelformat=empty} 
\caption{} 
\label{fig:sample-scanner} 
\end{wrapfigure}

\marginpar{
\vspace{.7cm} 
\color{Gray} 
\textbf{Figure \ref{fig:sample-scanner}. Document scanner} 
The \emph{DSmobile600} sheet-feed portable document scanner from \emph{Brother Mobile Solutions Inc.}.
}

Connection of an external source of light can be intrusive to the movement of the scanner carriage unit in a flatbed scanner which may cause discrepancies in the scanned image. In contrast, a sheet-feed scanner allows integration with an external source of light without being intrusive to the scanner operation. Since the scanning unit of a sheet-feed scanner is stationary, its operation is not affected by connection to an external source of light. A sheet-feed mechanism is therefore preferred over a flatbed construction to form the basis of a multispectral document scanner. The size of a sheet-feed scanner is mainly determined by the shorter edge of the supported page size which makes it compact and portable as shown in Figure~\ref{fig:sample-scanner}.

A broadband source of light (e.g., incandescent or fluorescent) reflects the average response of a scene over a wide spectral range, and therefore achieves low spectral fidelity. A multispectral source produces light in narrow spectral bands, attaining high spectral fidelity. Light Emitting Diodes (LEDs) can provide such selectivity required in the spectral profile of a multispectral light source. Another favorable characteristic of LEDs is that they are highly efficient compared to other sources of light.

\subsection{Electrical Schematic}

\begin{figure}[ht]
  \centering
  \includegraphics[width=1\linewidth]{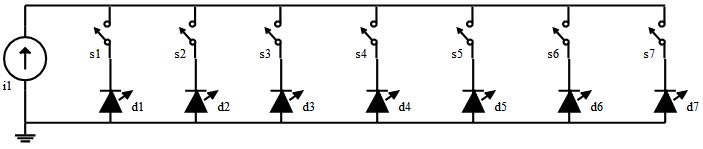}
  \caption{Schematic diagram of the multispectral light source. Connection layout of the LEDs (d1-d7) and the constant current source (i1) via switches (s1-s7).}\label{fig:elect-schematic}
\end{figure}

The electrical schematic of the multispectral light source is given in Figure~\ref{fig:elect-schematic}. It consists of a constant current source (i1) connected to narrow-band LEDs (d1-d7) via switch (s1-s7). The constant current source limits the current from surpassing the absolute maximum current rating of the LED. It also makes an LED glow with the same luminous power and spectral profile, making the system reliable. However, an inadvertent connection of multiple switches simultaneously can result in current being divided in several LEDs.

\begin{wrapfigure}[9]{l}{0.4\linewidth}
  \includegraphics[width=0.59\linewidth]{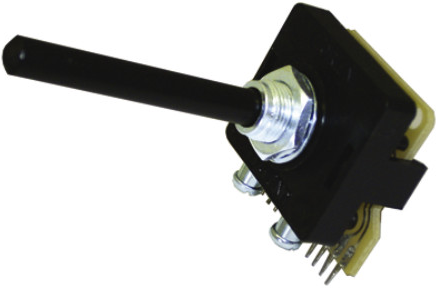}
  \includegraphics[width=0.39\linewidth]{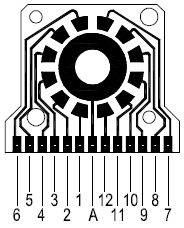}
\captionsetup{labelformat=empty} 
\caption{} 
\label{fig:PC-rotary} 
\end{wrapfigure}

\marginpar{
\vspace{.7cm} 
\color{Gray} 
\textbf{Figure \ref{fig:PC-rotary}. Rotary switch.} 
A 30 degree indexing, 12 way unipolar switch \emph{PT-6015} from \emph{Lorlin Electronics} and its diagram of terminal connections.
}

In order to ensure only one LED is powered at a time, a unipolar multi-way rotary switch is included in the design. It provides non-shorting, break-before-make contacts, to avoid overloading of the source with multiple LEDs during switching. It can handle high currents of 500mA @ 250Vac/dc which is within the operating limits of the constant current source (350mA). The switch and its terminal positions as viewed from the knob end of the spindle are shown in Figure~\ref{fig:PC-rotary}. Terminal A (middle) is connected to the positive end of the constant current source. Terminals 1-7 are connected to positive terminal of d1-d7, respectively. Terminals 8-12 are not utilized.

\begin{figure}[ht]
  \centering
  \subfigure[MicroPuck driver (ic1) powered by a low input voltage source (v1).]{\makebox[0.48\linewidth]{\includegraphics[width=0.3\linewidth]{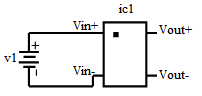}}} \hfill
  \subfigure[Assembled low input voltage - constant current source]{\makebox[0.45\linewidth]{\includegraphics[width=0.3\linewidth]{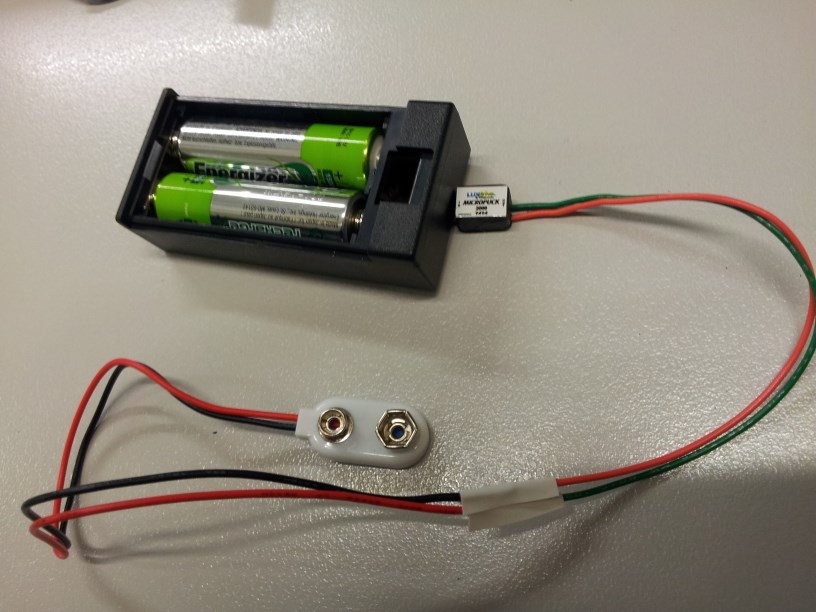}}}
   \subfigure[BuckPuck driver(ic2) powered by a high input voltage source (v2).]{\makebox[0.48\linewidth][c]{\includegraphics[width=0.28\linewidth]{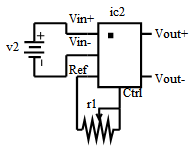}}}  \hfill
  \subfigure[Asssembled high input voltage - constant current source]{\makebox[0.45\linewidth]{\includegraphics[width=0.3\linewidth]{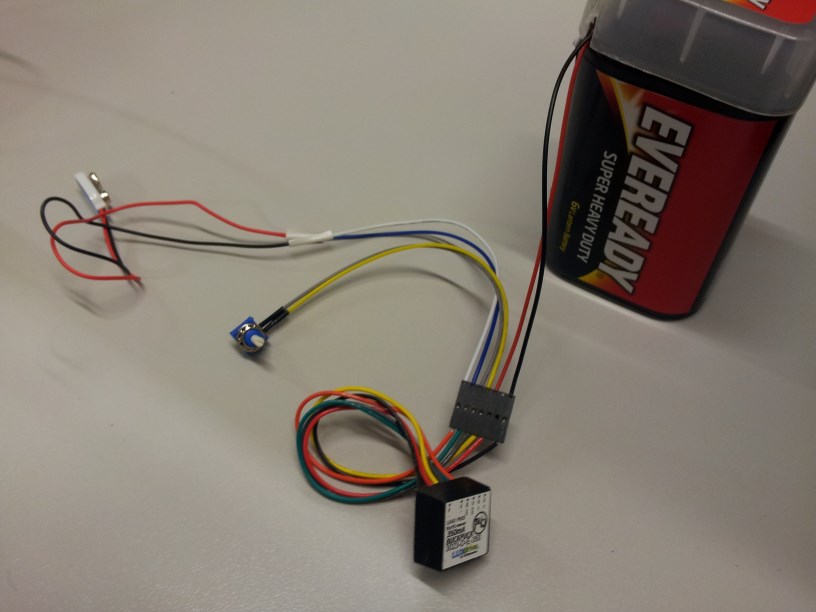}}}
  \caption{Constant current sources (a) Schematic of the low input voltage source. The input terminals (V$_\textrm{in+}$,V$_\textrm{in-}$) are connected to a low voltage source (v1=0.8-3 Vdc) and the output terminals (V$_\textrm{out+}$,V$_\textrm{out-}$) are connected to LED. (b) Schematic of the high input voltage source. The input terminals (V$_\textrm{in+}$,V$_\textrm{in-}$) are connected to a high voltage source (v1=7-32 Vdc) and the output terminals (V$_\textrm{out+}$,V$_\textrm{out-}$) are connected to LED. The potentiometer (r1) allows dimming control (Ctrl) via internal reference (Ref).}
  \label{fig:cc-source}
\end{figure}

Two different constant current sources were designed depending on availability of a low or high input voltage source. The electrical schematic of the sources and their assembled form are shown in Figure~\ref{fig:cc-source}.

\subsubsection{Low Input Voltage - Constant Current Source}
The low input voltage constant current source uses a MicroPuck LED Power Module which can provide a constant (350mA) current to a single LED. The miniature design allows flexibility to use one or two AA sized batteries to power the module. It provides maximum current to the LEDs while mimicking the light drop-off of an incandescent bulb, which dims as the batteries drain. However, the current drops only at very low voltages, allowing maximum operational time. The driver has two input pins (V$_\textrm{in+}$, V$_\textrm{in-}$) and two output pins (V$_\textrm{out+}$, V$_\textrm{out-}$).

\subsubsection{High Input Voltage - Constant Current Source}
The high input voltage constant current source uses a BuckPuck LED Power Module which can provide a constant (350mA) current to multiple LEDs, although the proposed switch ensures only one LED is powered at a time. The module provides manual dimming control through a potentiometer which uses internal reference from the BuckPuck driver. It also has built-in open-circuit and short-circuit protection. The module has four input pins (V$_\textrm{in+}$, V$_\textrm{in-}$, Ref, Ctrl) and two output pins (V$_\textrm{out+}$, V$_\textrm{out-}$).

\subsection{Spectral Profile}

\begin{figure}[ht]
  \centering
  \subfigure{\label{fig:sample-LEDs}\includegraphics[width=0.23\linewidth]{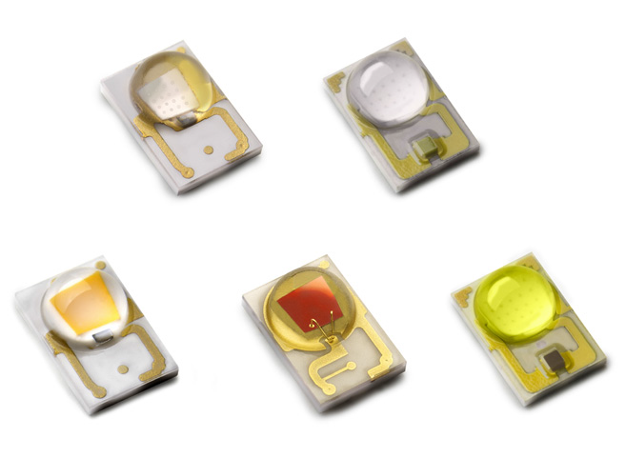}}
  \subfigure{\label{fig:SPD}\includegraphics[width=0.74\linewidth]{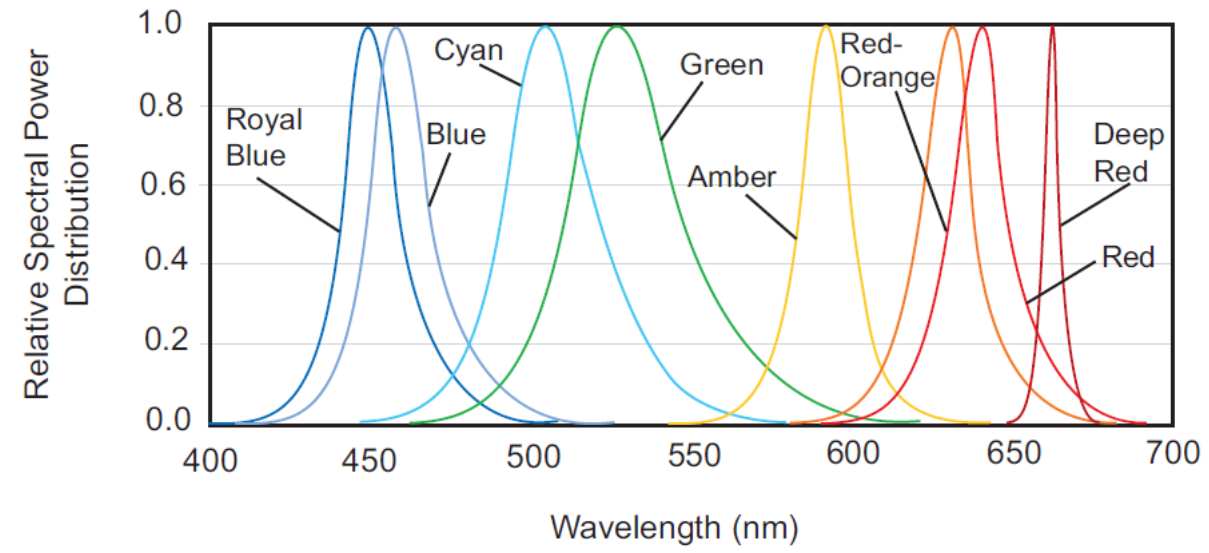}}
  \caption{(a) A few \emph{Philips Luxeon Rebel} LEDs in different colors. (b) Relative Spectral Power Distribution of the LEDs} \label{fig:spectral-design}
\end{figure}

The choice of colored LEDs is important for description of the spectral profile of the multispectral light source. The spectral characteristics of LEDs are characterized by two main parameters, i.e. the center wavelength and the spectral bandwidth. The relative spectral power distribution of the LEDs is given in Figure~\ref{fig:SPD}. These LEDs almost cover the range of visible electromagnetic spectrum (400nm-700nm) at approximately regular intervals.

The spectral parameters of the LEDs are provided in Table~\ref{tab:LED-specs}. Note that the LEDs are spread across the spectrum with sufficiently narrow-bands and high luminous power which makes an effective multispectral light source.

\begin{table}[ht] 
  \centering
  \caption{Specifications of Luxeon Rebel LED series used in multispectral light source.}
  \label{tab:LED-specs}
  {\footnotesize
  \begin{tabular}{|l|c|c|c|l|}
     \hline \hline
    \multirow{2}{*}{Color} & \multirow{2}{*}{Wavelength} & \multirow{2}{*}{Bandwidth} & \multirow{2}{*}{Luminous Flux} &\multirow{2}{*}{Part Number}   \\
    & & & &  \\ \cline{2-5}
                    & (nm)  &  (nm)  & (lm,mW)&  model          \\ \hline \hline
    Deep Red        & 655   &   20   & 640    & LXM3-PD01       \\
    Red-Orange      & 617   &   20   & 90     & LXML-PH01-0050  \\
    Amber           & 590   &   20   & 77     & LXML-PL01-0040  \\
    Green           & 530   &   30   & 150    & LXML-PM01-0090  \\
    Cyan            & 505   &   30   & 122    & LXML-PE01-0070  \\
    Royal Blue      & 447.5 &   20   & 1030   & LXML-PR02-A900  \\
     \hline
   \end{tabular}}
\end{table}

\subsection{Optical Configuration}

The purpose of optical assembly is to transmit multispectral light into the to scanner light guide. The concentrator optics are suitable for beam insertion into fiber optic bundles or light guides. Two different optical arrangements were proposed for multispectral light source shown in Figure~\ref{fig:struct-design}.

\subsubsection{Linear LEDs with concentrator lens}

In this arrangement, an LED is pre-soldered to a base with anode(+) and cathode(-) connections at locations as shown in Figure~\ref{fig:single-assembly}. Multiple such units, each with a different color LED together make a multispectral light source. A fiber beam lens (\emph{Carclo Inc.}) focuses light from LED into an 8 degree narrow beam at a focal distance of 11mm. The diameter of the lens is 20mm and conforms to the LED base. It requires a circular lens holder shown in Figure~\ref{fig:carclo-lens} which is affixed to the base using a double-sided tape. The holder positions the lens at appropriate distance from the LED for maximum luminous transmission. The focused beam is then fed to a flexible light guide.

\subsubsection{Array LED base with cluster concentrator optic}

\begin{figure}[ht]
\centering
  \begin{minipage}[b]{0.35\linewidth}
  \centering
  \subfigure[Single LED assembly]{\label{fig:single-assembly}\makebox[1\linewidth]{\includegraphics[width=0.3\linewidth]{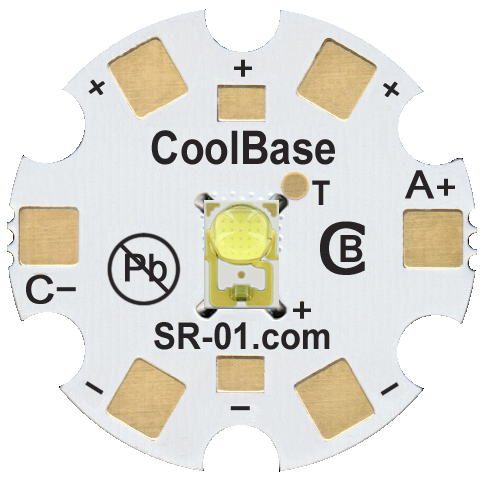}}}
  \subfigure[7-LED round assembly]{\label{fig:array-assembly}\makebox[1\linewidth]{\includegraphics[width=0.6\linewidth]{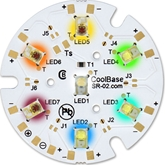}}}\\
  \end{minipage} \hfill
  \begin{minipage}[b]{0.63\linewidth}
  \centering
  \subfigure[Fiber coupler concentrator lens and round optic holder]{\label{fig:carclo-lens}\makebox[1\linewidth]{\includegraphics[width=0.15\linewidth]{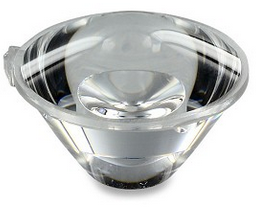}
  \includegraphics[width=0.15\linewidth]{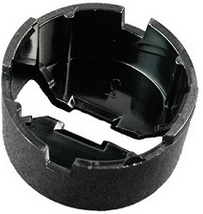}}}\\ [29pt]
  \subfigure[Multi-cell cluster concentrator optic with pegged feet]{\label{fig:cluster-optic}\makebox[1\linewidth]{\includegraphics[width=0.3\linewidth]{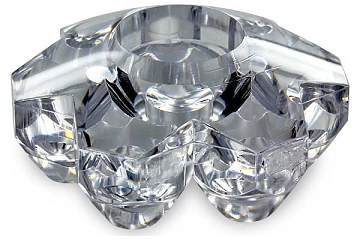}}}
  \end{minipage}
  \caption{Different components in the two optical configurations of multispectral light source.}\label{fig:struct-design}
\end{figure}

In this arrangement, an array of LEDs is pre-soldered to a single base with separate anode(+) and cathode(-) connections for each unit at locations as shown in Figure~\ref{fig:array-assembly}. A cluster concentrator optic (\emph{Polymer Optics Ltd.}) focuses light from seven LEDs into a 12mm narrow beam at a focal distance of 25mm. It uses an optical grade poly-carbonate material for thermal stability and system durability which results in high light collection efficiency (85\%). The use of array LEDs and cluster optic makes the light source compact and rigid.

\begin{figure}[ht]
\centering
  \subfigure[Single LED assembly with soldered connections]{\makebox[0.33\linewidth]{\includegraphics[width=0.15\linewidth]{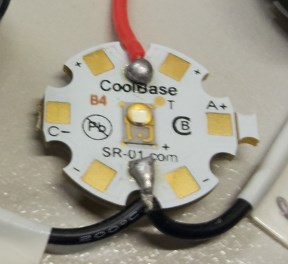}}}
  \subfigure[Carclo lens in round lens holder]{\makebox[0.33\linewidth]{\includegraphics[width=0.15\linewidth]{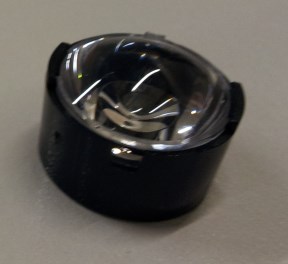}}}
  \subfigure[Patched up components of linear optical configuration]{\makebox[0.3\linewidth]{\includegraphics[width=0.25\linewidth]{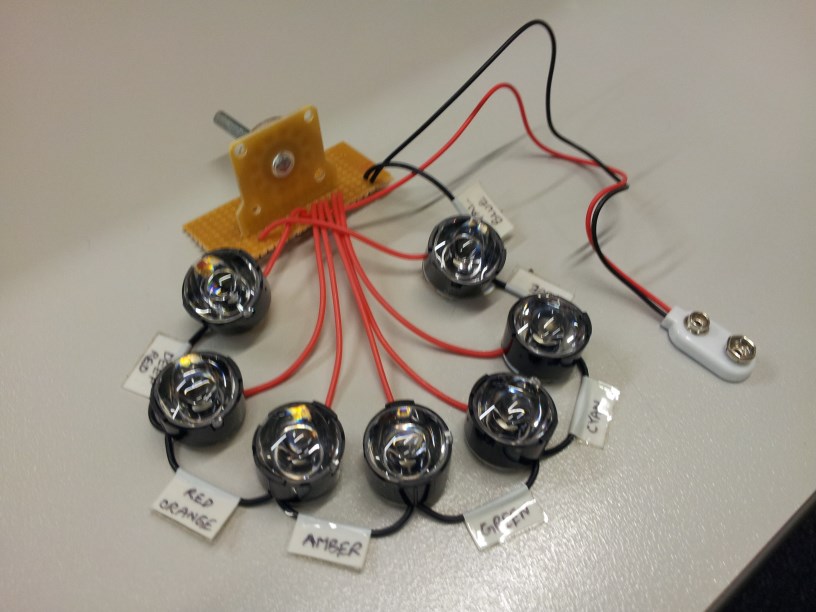}}}\\
  \subfigure[Array LED assembly with soldered connections]{\makebox[0.33\linewidth]{\includegraphics[width=0.25\linewidth]{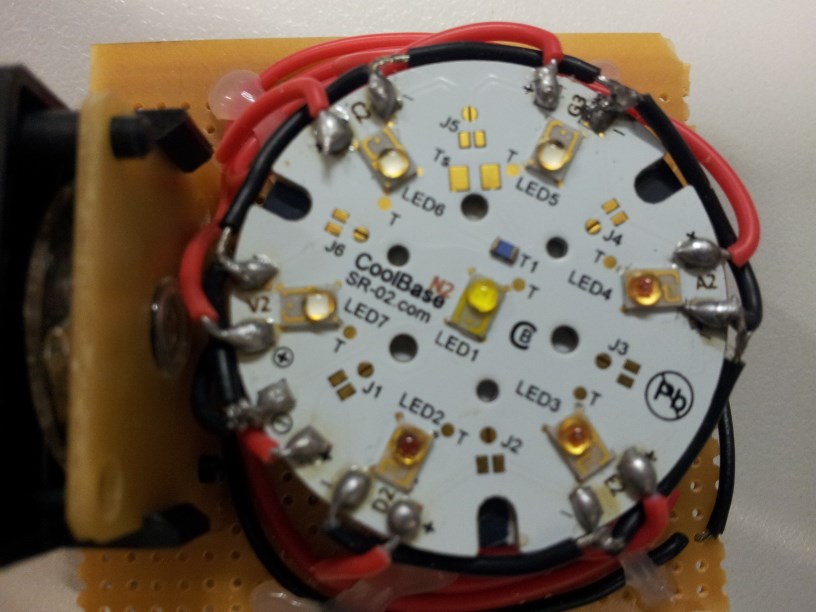}}}
  \subfigure[Array LED assembly with cluster concentrator lens]{\makebox[0.33\linewidth]{\includegraphics[width=0.25\linewidth]{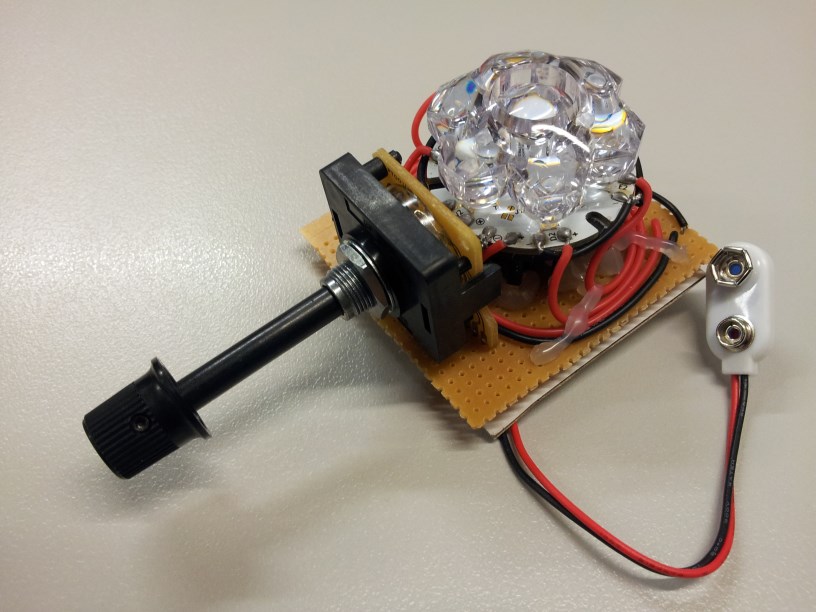}}}
  \subfigure[Patched up components of array optical configuration]{\makebox[0.3\linewidth]{\includegraphics[width=0.25\linewidth]{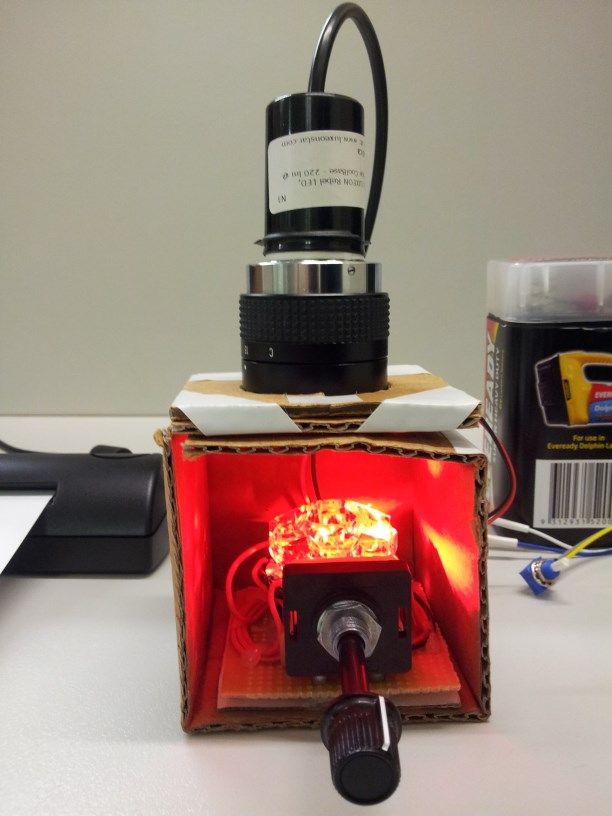}}}
  \caption{Steps for assembly of different optical configurations, linear (top row) and array (bottom row)}\label{fig:optical-mod}
\end{figure}

The two optical configurations after assembly are shown in Figure~\ref{fig:optical-mod}.

\subsection{Thermal Consideration}

\begin{wrapfigure}[7]{l}{0.15\linewidth}
\includegraphics[width=1\linewidth]{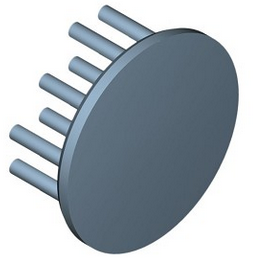}
\captionsetup{labelformat=empty} 
\caption{} 
\label{fig:heat-sink} 
\end{wrapfigure}

\marginpar{
\vspace{.7cm} 
\color{Gray} 
\textbf{Figure \ref{fig:heat-sink}. Heat sink} 
A 3D view of heat sink \emph{CN40-15B} from \emph{ALPHA Co. Ltd}.
}

The use of high-power LEDs introduces considerable thermal efficiency issues which can overheat the LEDs if not catered for correctly. A heat sink is an affordable device for maintaining near constant temperature of LEDs for long periods of operation. We used a natural convection heat sink, which is a 40mm round base with 15mm high legs as shown in Figure~\ref{fig:heat-sink}. It has the highest thermal efficiency (degree Celsius per Watt) in the CN40 series of heat sinks.

\subsection{Scanner Modification}

Modification of the sheet-feed scanner consists of the following steps:

\begin{enumerate}
\item Gaining access to the internal micro illumination source (RGB LED) of the scanner.
\item Removing/Disabling of the internal RGB LED.
\item Structural modification of scanner housing for placement of flexible light guide.
\item Connection of external light source to the internal light guide via flexible light guide.
\end{enumerate}

\begin{figure}[ht]
\centering
  \subfigure[Gain access by removing top cover]{\includegraphics[width=0.32\linewidth]{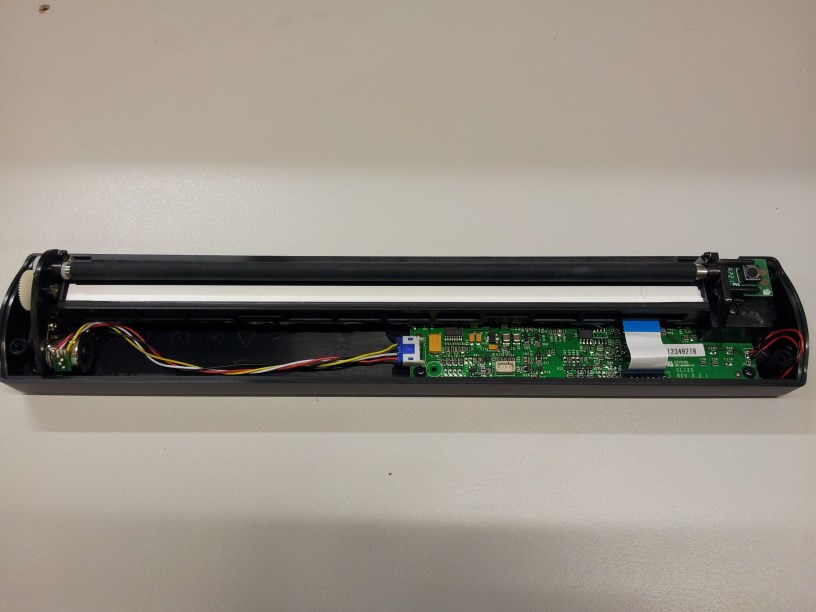}}
  \subfigure[Release components from hinge support]{\includegraphics[width=0.32\linewidth]{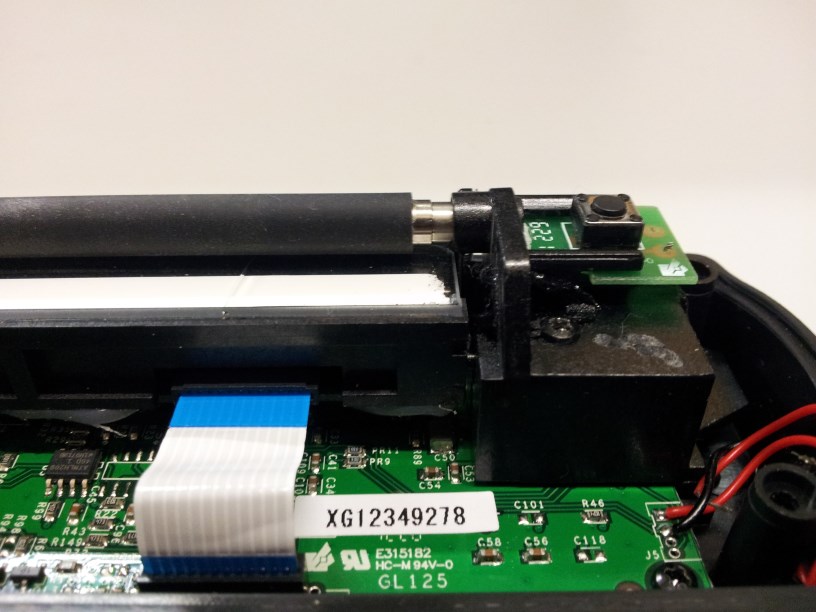}}
  \subfigure[Disengage RGB LED from scanner sensor]{\includegraphics[width=0.32\linewidth]{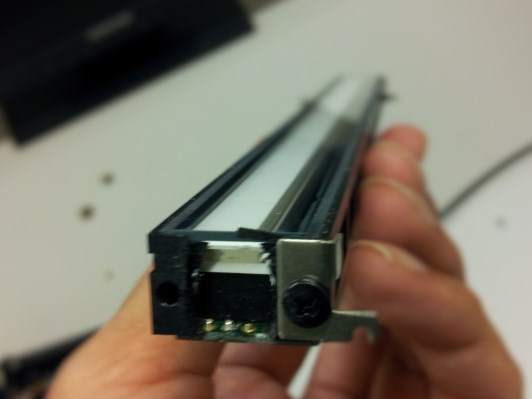}}\\
  \subfigure[Make provision for flexible light guide]{\includegraphics[width=0.32\linewidth]{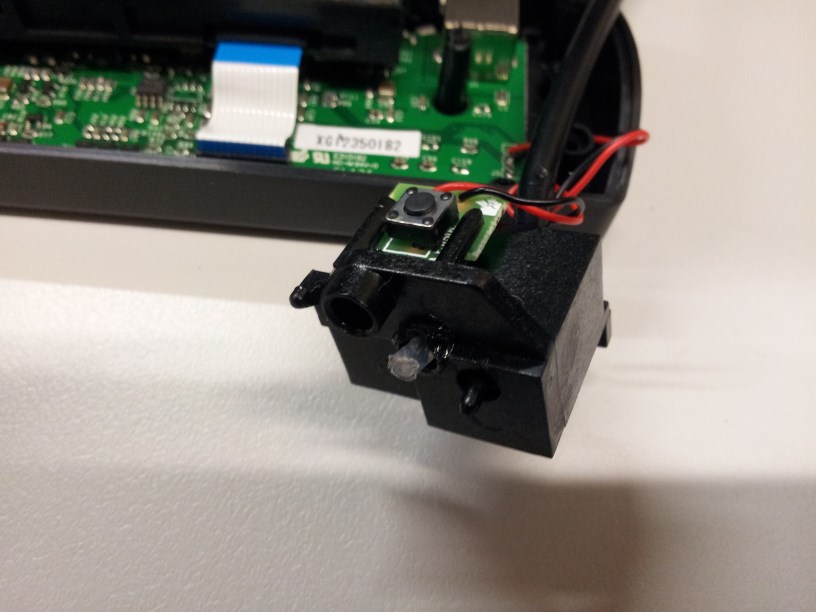}}
  \subfigure[Reinstall components]{\includegraphics[width=0.32\linewidth]{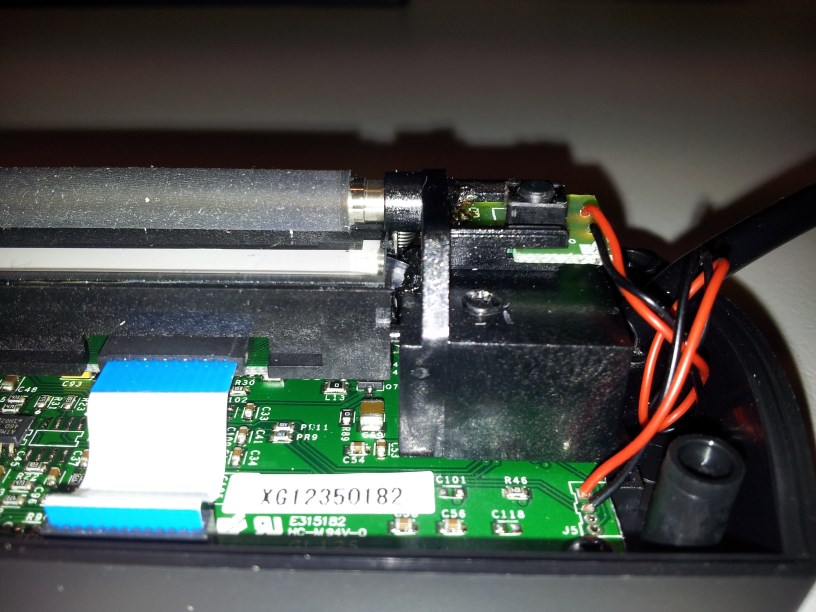}}
  \subfigure[Replace top cover]{\includegraphics[width=0.32\linewidth]{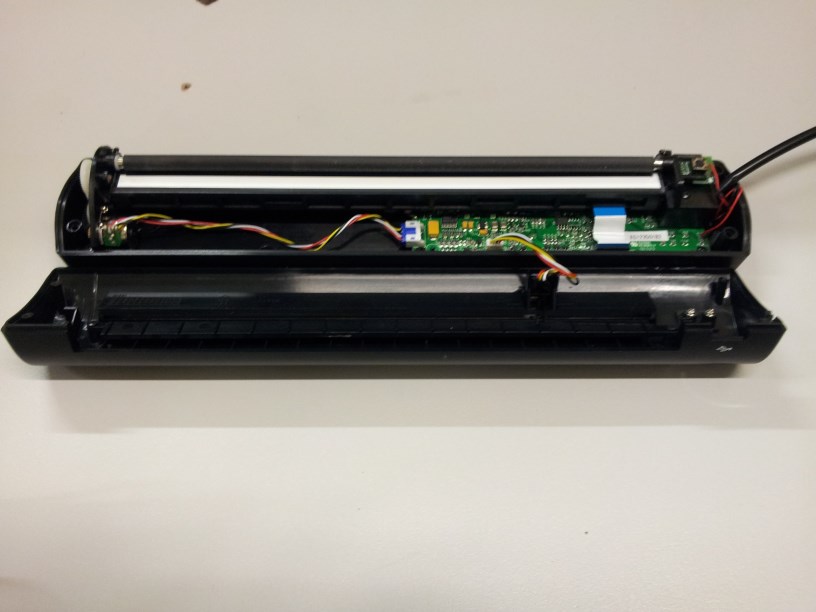}}\\
  \subfigure[Connection to multispectral light source]{\includegraphics[width=0.32\linewidth]{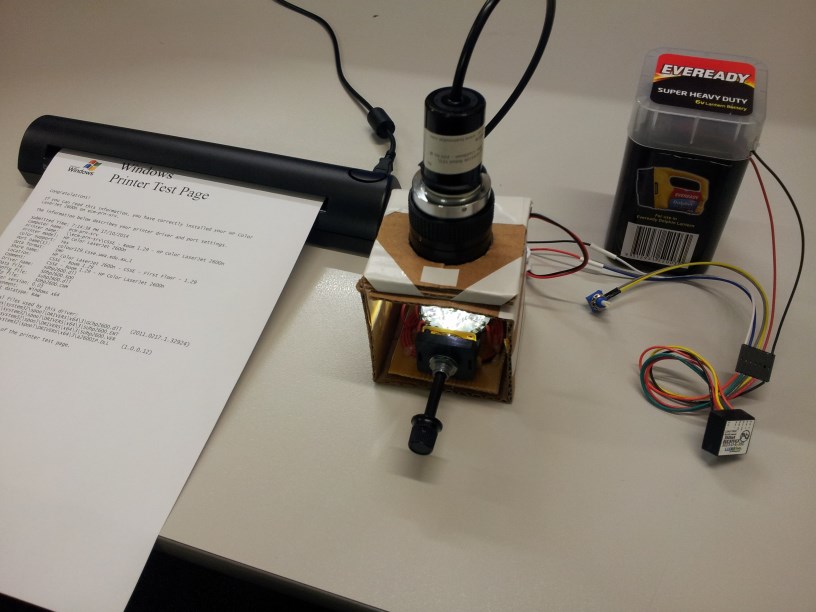}}
  \caption{Steps of scanner modification and assembly of multispectral light source.}\label{fig:scanner-mod}
\end{figure}

The procedure is illustrated in detail in Figure~\ref{fig:scanner-mod}

\section{Multispectral Document Scanning}

\begin{figure}[ht]
  \centering
\fbox{\includegraphics[width=0.12\linewidth]{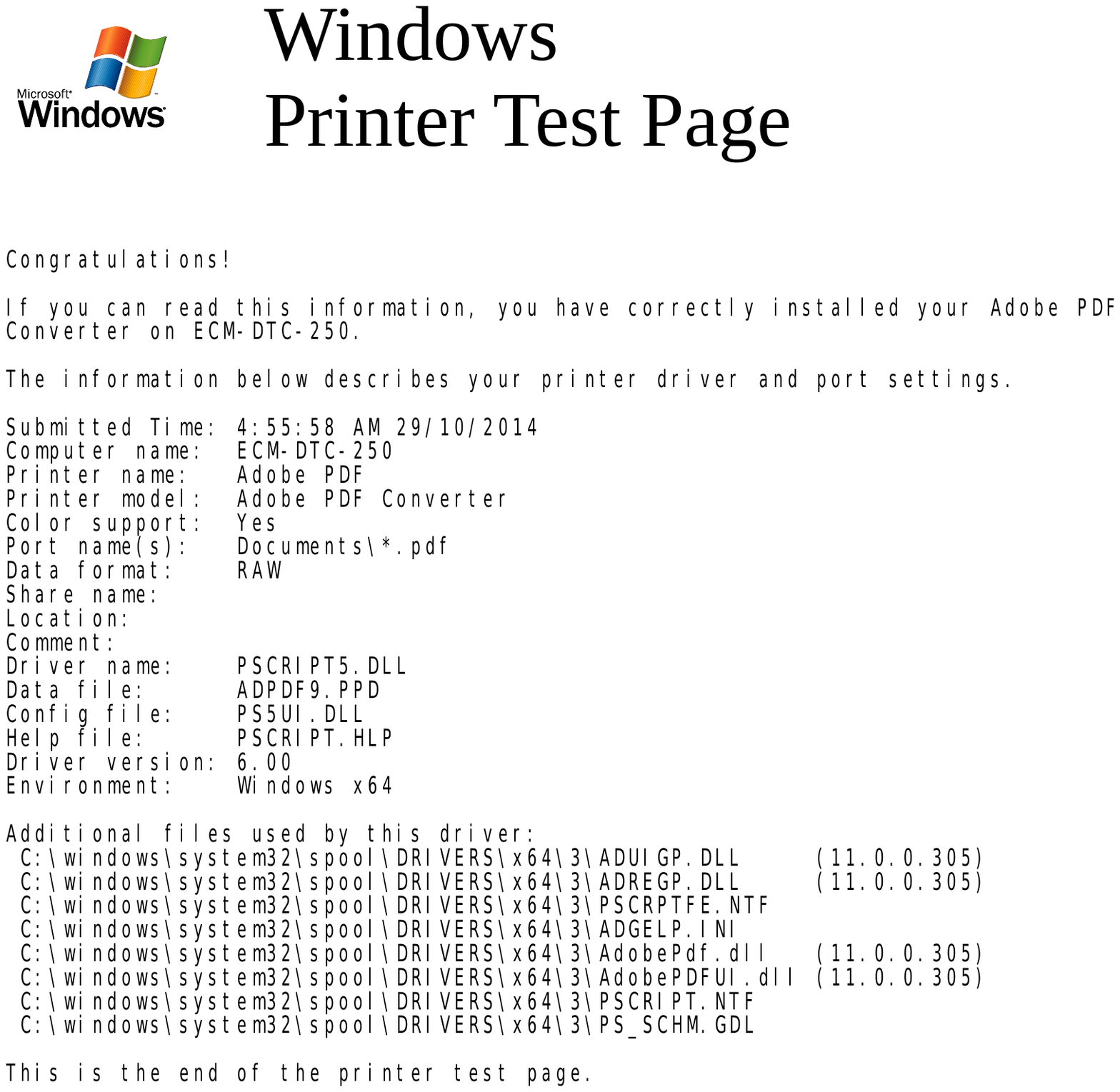}}
\includegraphics[width=0.85\linewidth]{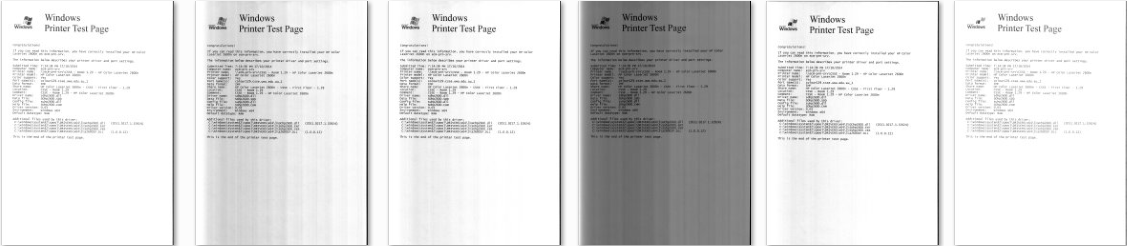}\\ [5pt]
\begin{minipage}[b]{0.1\linewidth}
\centering
\includegraphics[trim = 1.25cm 25.51cm 16.8cm 1.48cm, clip, width=1\linewidth]{Images/TestPage}
\end{minipage} \hspace{1pt}
\includegraphics[trim = 55 140 38 120, clip, width=0.85\linewidth]{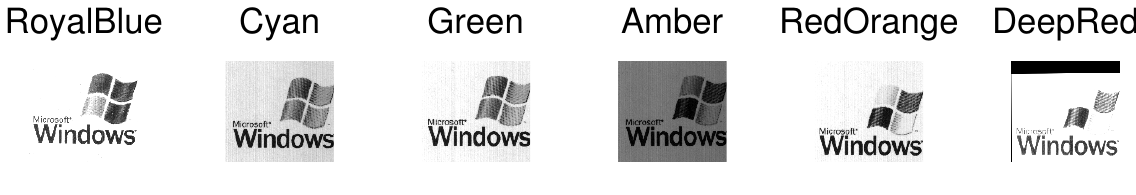}
  \caption{Multispectral image of a logo from printer test document. Note how the intensity of portions of the logo changes according to spectral bands.}\label{fig:logos}
\end{figure}

In order to test the multispectral document scanner, a test page was printed from an HP Laserjet Color printer. The RGB true color and various bands of a logo in the test page captured by the multispectral scanner are shown in Figure~\ref{fig:logos}. The logo has red, orange, green and blue elements and the text at the bottom. Notice that the colored portions of the logo have a characteristic response to the multispectral light. This demonstrates the ability of the scanner to capture fine detail in spectral bands.

The recent availability of high-resolution printers, has not only supported useful purposes, but also opened ways for illegal use in forgery of documents. This has consequently persuaded printer manufacturers to hide an invisible counterfeit protection code which holds information for printer identification. This unique code is printed in every document, in the form of repeated pattern of yellow dots. The unique pattern can in turn be used to identify the source of a document. The multispectral document scanner successfully captures this unique dot-pattern.

\begin{figure}[ht]
\centering
\subfigure[Royal Blue band]{\includegraphics[width=0.33\linewidth]{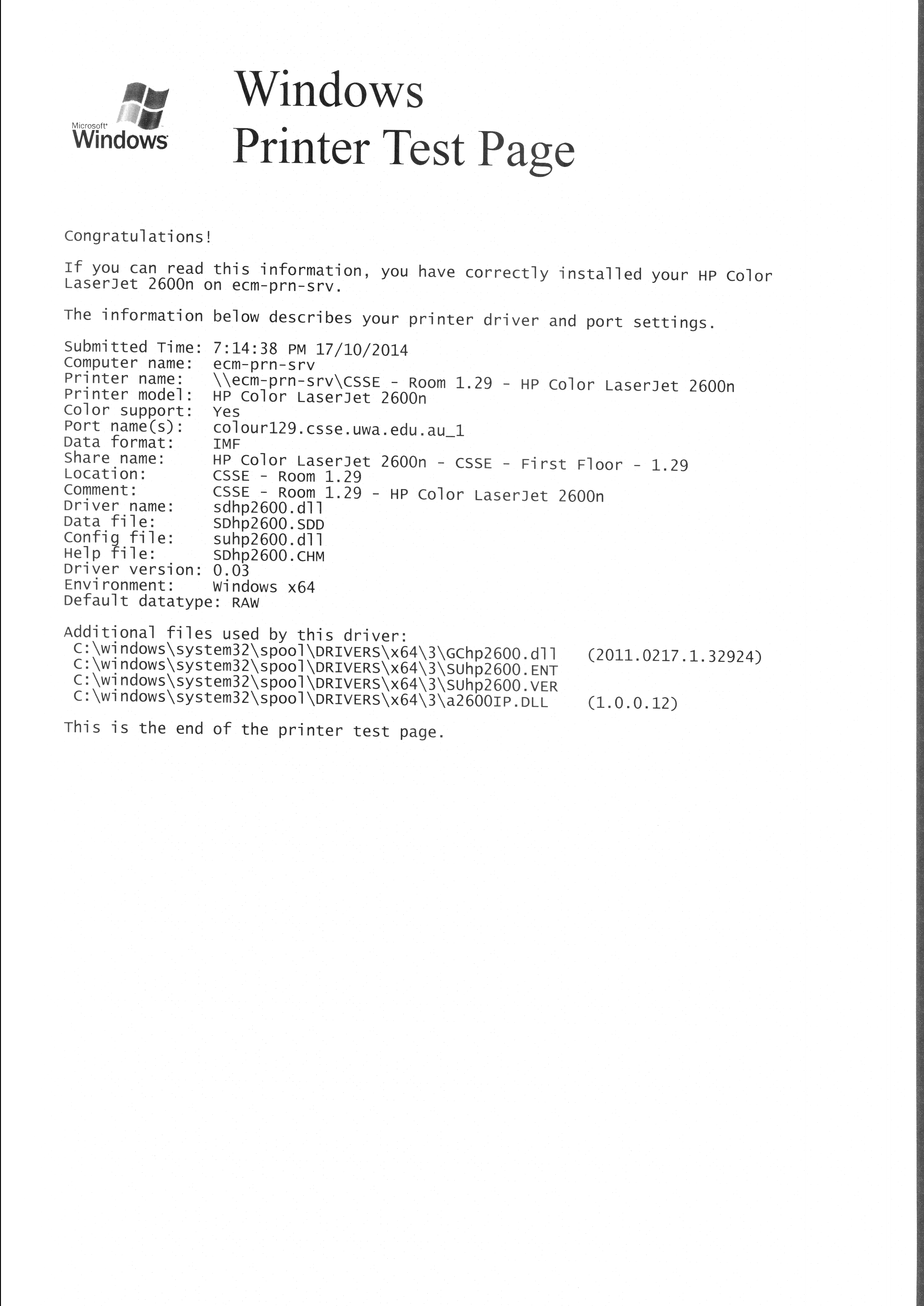}}
\subfigure[Processed image]{\includegraphics[width=0.64\linewidth]{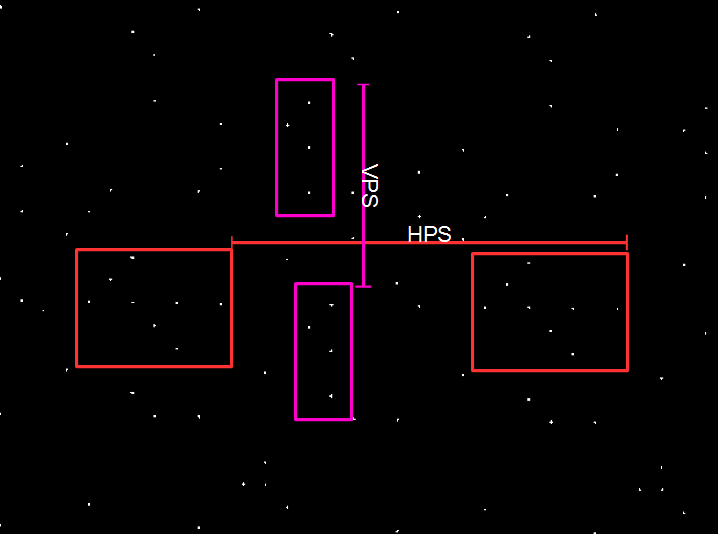}}  \caption{The Royal blue band of a test printer document (zoom to 6400\% to clearly view the pattern). Counterfeit protection pattern extracted by processing of the Royal Blue band. The coded patterns are separated by HPS and VPS.}\label{fig:pattern}
\end{figure}

The unique patterns of different printers can be identified in terms of geometrical relationships. Two important parameters that form these relationships are the Horizontal Pattern Separation (HPS) distance and the Vertical Pattern Separation (VPS) distance. A raw image of the Royal Blue band of the scanned test page is shown in Figure~\ref{fig:pattern}.

\section{Conclusion}
We presented the design of a prototype multispectral document scanner which is demonstarted to capture subtle features in a document using multispectral light source. The multispectral light source was designed to cover the full range of visible electromagnetic spectrum and connected to a portable sheet-feed document scanner. The prototype design has potential to be transformed into a fully functional device suitable for portable document analysis.

\section*{Acknowledgments}
This research work was partially funded by the ARC Grant DP110102399 and the UWA Grant 00609~10300067.


\bibliographystyle{abbrv}

\bibliography{main}

\end{document}